\documentclass{article}
\usepackage{spconf,amsmath,epsfig}
\usepackage{amssymb}
\usepackage{amsthm}
\usepackage{algorithm}
\usepackage{algpseudocode}
\usepackage{tikz}
\usepackage{graphicx}
\usepackage{subfig}
\usepackage{epstopdf}
\usepackage{epsfig}
\usetikzlibrary{tikzmark,calc}
\usepackage{caption}
\usepackage{float}
\usepackage{multirow}
\usepackage{todonotes}
\usepackage{microtype}

\numberwithin{equation}{section} 

\def\R{{\mathbb{R}}}

\DeclareMathOperator*{\argmin}{arg\,min}
\DeclareMathOperator*{\argmax}{arg\,max}

\renewcommand{\mod}{\mathrm{mod}}

\title{Reconstruction from Periodic Nonlinearities, \\ With Applications to HDR Imaging}
\name{Viraj Shah, Mohammadreza Soltani, Chinmay Hegde\thanks{This work was supported in part by grants from the National Science Foundation (NSF CCF-1566281) and NVIDIA.}
}
\address{ECpE Department, Iowa State University, Ames, IA, 50010}
 
\begin{document}
\ninept
\maketitle
\begin{abstract}
We consider the problem of reconstructing signals and images from periodic nonlinearities. For such problems, we design a measurement scheme that supports efficient reconstruction; moreover, our method can be adapted to extend to compressive sensing-based signal and image acquisition systems. Our techniques can be potentially useful for reducing the measurement complexity of high dynamic range (HDR) imaging systems, with little loss in reconstruction quality. Several numerical experiments on real data demonstrate the effectiveness of our approach.
\end{abstract}
%


\section{Introduction}
\label{sec:intro}

\subsection{Motivation}

Reliable estimation of a signal (or image) from nonlinear observations is of fundamental interest to several signal processing and machine learning applications. However, such an estimation is confounded by cases where the nonlinearity in each observation is well-modeled by a \emph{periodic} function such as a sinusoidal function, or sawtooth function, or a square-wave function. Periodic functions are many-to-one mappings, and inverting them can be challenging.

Our focus in this paper is a special kind of periodic nonlinear observation model encountered in high-dynamic range (HDR) imaging. It is well known that real world scenes contain a large range of brightness levels. However, due to hardware limitations, not all brightness levels can be accurately captured using conventional photography; if tuned incorrectly, most scene intensity levels can lie in the saturation region of the image sensors, causing loss of scene information. Similar problems arise in the case of multiplexed imaging systems, such as lensless and coded aperture imaging~\cite{codedaperture,asif2017flatcam}.

One solution is to increase the dynamic range of the image sensors, but this can lead to expensive hardware. An alternative solution is to deploy a special type of image sensor that {wraps} the observed intensity value at a pixel over a given dynamic range. This is analogous to the familiar \emph{modulo} operation with respect to a parameter $R$, and we call this stylized imaging system a \emph{modulo camera}~\cite{ICCP15_Zhao}.  
Fig.~\ref{fig:func}(a) (black) depicts the modulo nonlinearity, and a major challenge is to undo the effect of this transformation for each observed pixel.

An added challenge in HDR imaging arises due to \emph{quantization}. In fact, the ``true" observations in a modulo camera are quantized versions of the (idealized) modulo observation, and the errors caused in the quantization propagates into the estimation process. Loss of information in the quantization process is unavoidable in principle, and the effect of quantization is magnified with fewer quantization levels. In acquisition systems with low bit-depth, such estimation errors can be very pronounced. Fig.~\ref{fig:func}(a) (cyan) depicts the quantization nonlinearity incurred during the observation process.

\subsection{Setup}

\begin{figure}[t]
	
	\begin{center}
		\begingroup
		\setlength{\tabcolsep}{0.1pt} 
		\renewcommand{\arraystretch}{.1} 
		\begin{tabular}{ccc}      
			\multicolumn{3}{c}{\begin{tikzpicture}
				\draw[<->,thick] (-3,0)--(3,0) node[anchor=north]{$t$};
				\draw (0,0) node[anchor=north]{$0$};
				\draw (0,1.1) node[anchor=west] {$R$};
				\draw (1,0) node[anchor=north]{$R$};
				\draw (2,0) node[anchor=north] {$2R$};
					\draw (-1,0) node[anchor=north]{$-R$};
				\draw (-2,0) node[anchor=north] {$-2R$};
				\draw[] (2,2) node[anchor=west] {{$Qof(t)$}};
				\draw[cyan,thick] (1.6,2) -- (2,2);
				\draw[->,thick] (0,0)--(0,2);
				\draw[] (2,1.5) node[anchor=west] {{$f(t)$}};
				\draw[thick] (1.6,1.5) -- (2,1.5);
				\draw[thick] (-2,0) --(-1,1)-| (-1,0) -- (0,1) -| (0,0) --(1,1)-| (1,0) -- (2,1) -| (2,0);
				\draw[densely dotted,thick] (2,0)--(2.5,0.5);
				\draw[densely dotted,thick] (-2,0)|-(-2,1) -- (-2.5,0.5);
				\draw[thick, cyan] (-2,0) -- ++(0.5,0)-| ++(0,0.5) -- ++(0.5,0) -| ++(0,-0.5) -- ++(0.5,0)-| ++(0,0.5) -- ++(0.5,0) -| ++(0,-0.5) -- ++(0.5,0)-| ++(0,0.5) -- ++(0.5,0) -| ++(0,-0.5) -- ++(0.5,0)-| ++(0,0.5) -- ++(0.5,0) -| ++(0,-0.5);
				\end{tikzpicture}}\\
			\multicolumn{3}{c}{(a)}\\
			\includegraphics[trim = 10mm 60mm 25mm 40mm,clip, width = 0.32\linewidth]{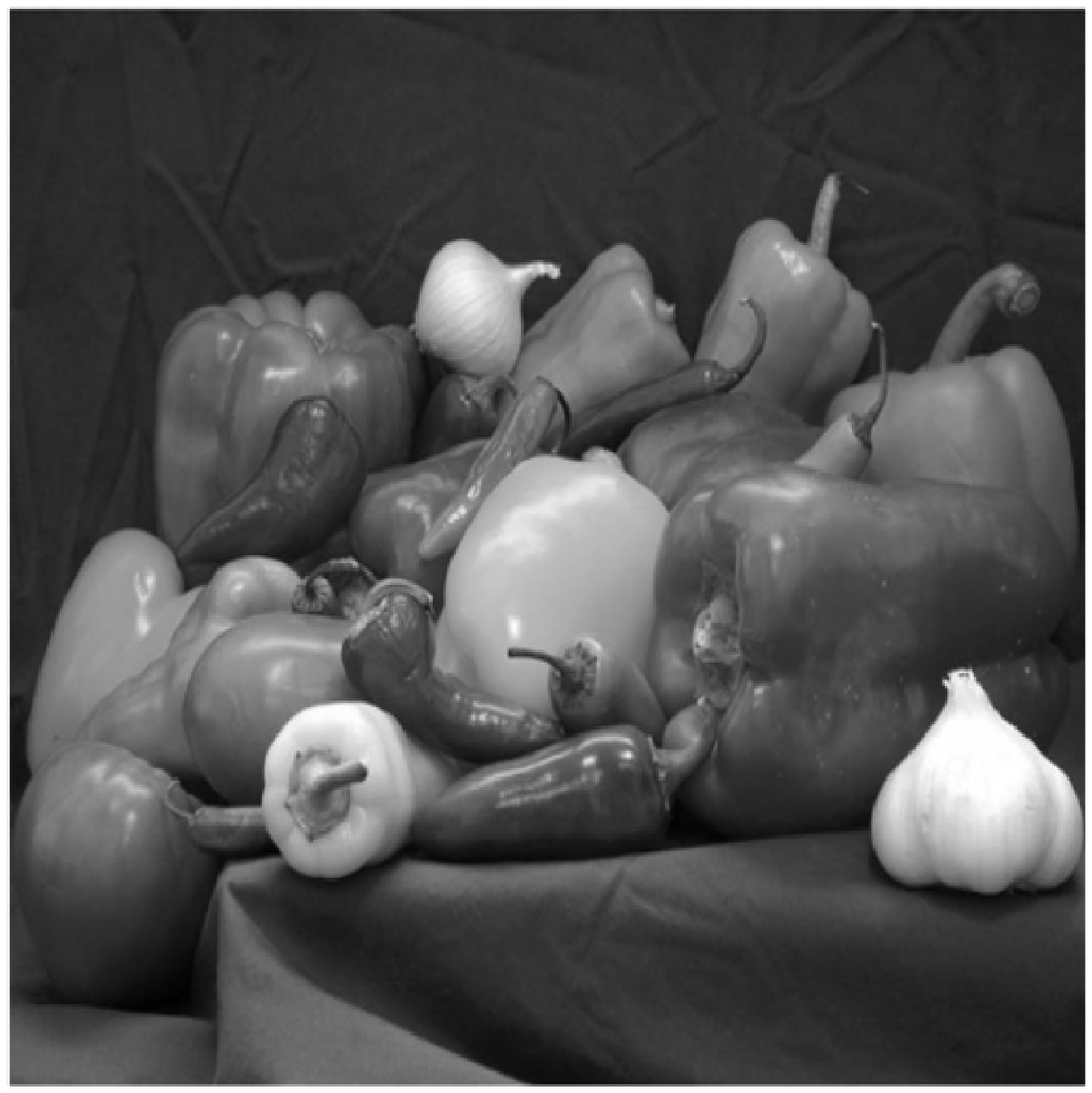}&
			\includegraphics[trim = 10mm 60mm 25mm 40mm,clip, width = 0.32\linewidth]{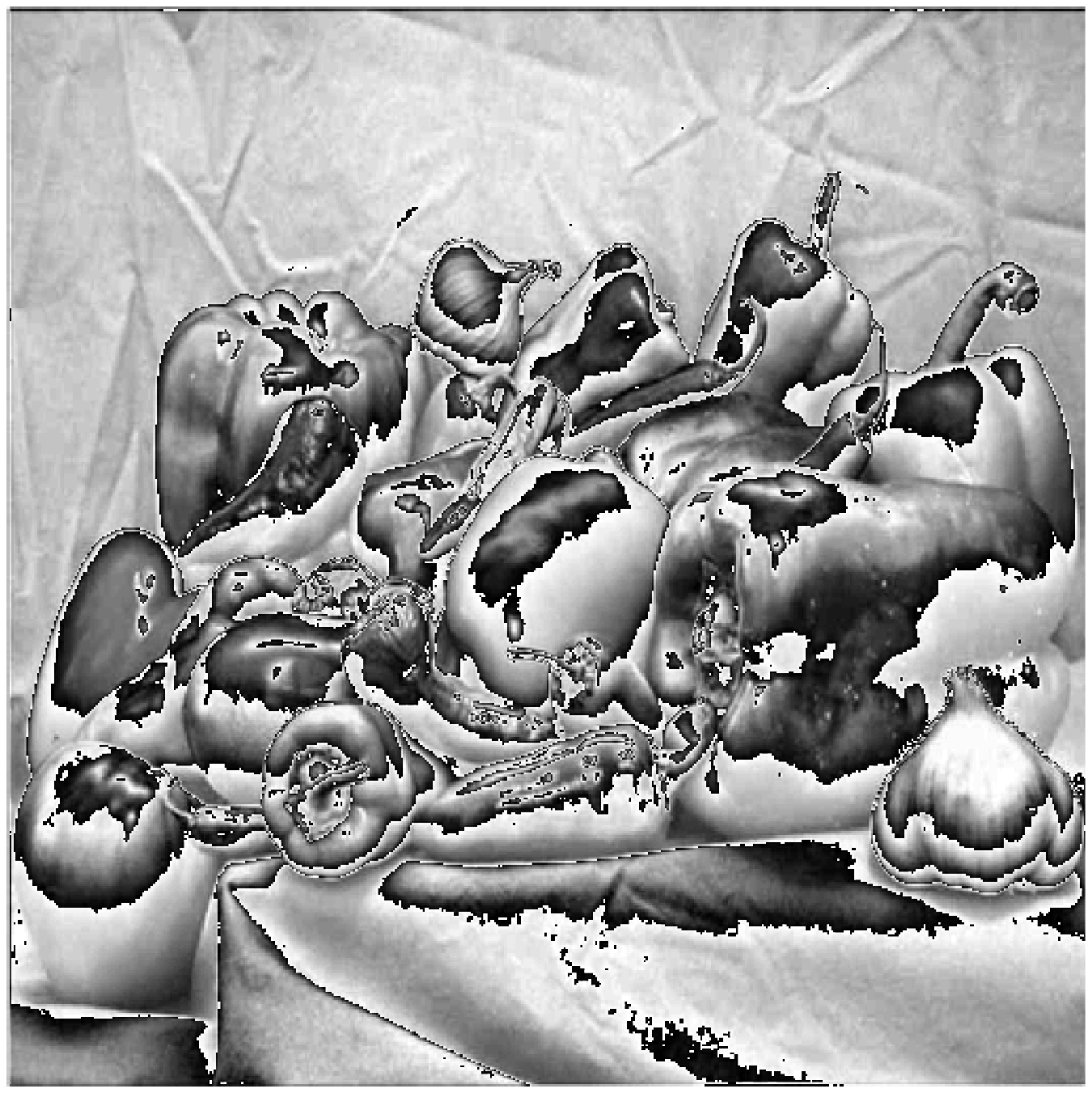}&
			\includegraphics[trim = 10mm 60mm 25mm 40mm,clip,width = 0.32\linewidth]{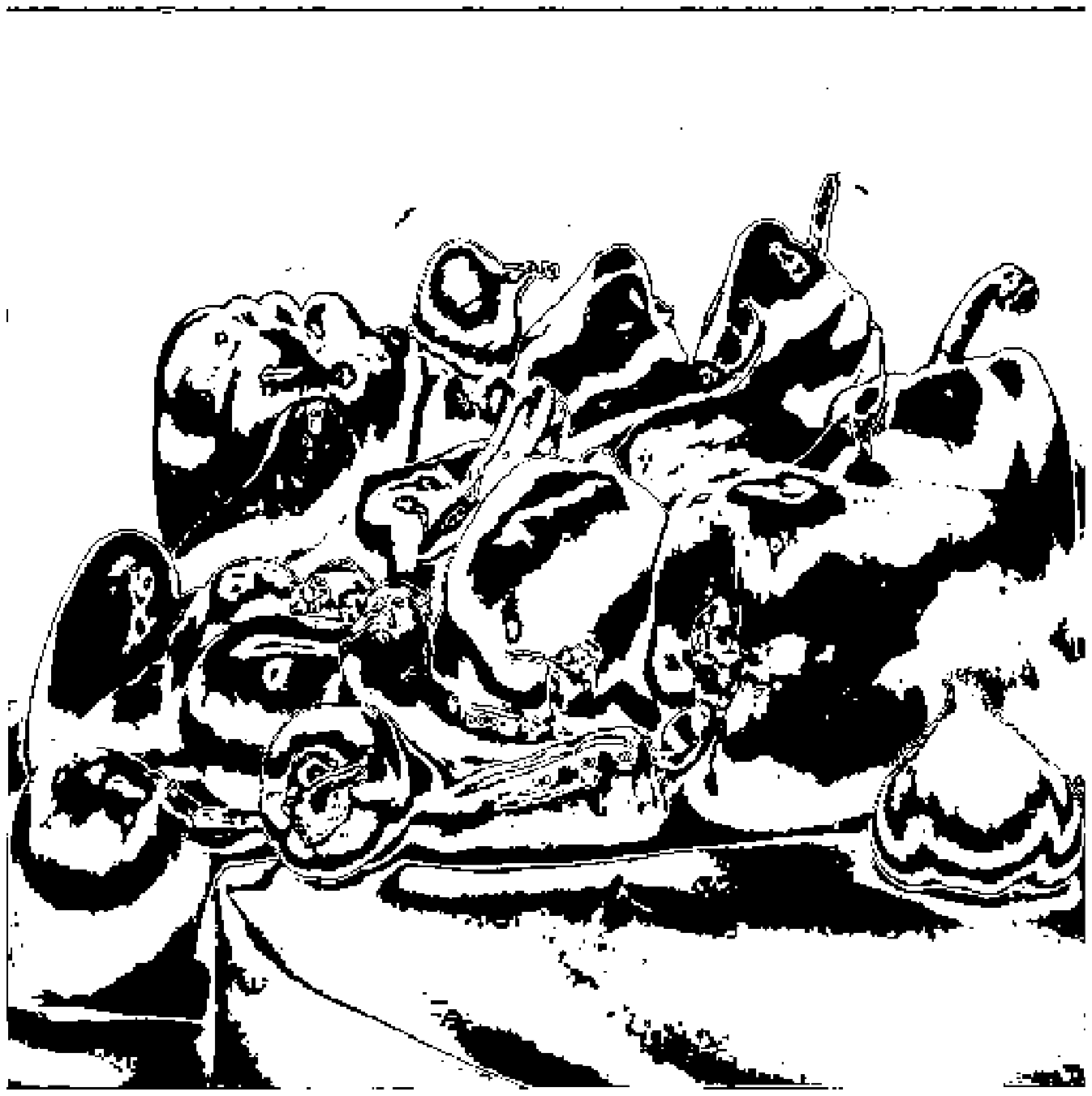} \\
			(b) & (c) & (d)
		\end{tabular}
		\endgroup
	\end{center}
	\caption{\small{\emph{ (a) Modulo function, $f(t) = \mod(t,R)$ and quantized modulo function, $Qof(t)$; (b,c,d) Depiction of forward model. An input image (b) is transformed via a modulo function $f(t) = \mod(t,R)$, to (c). Such a ``modulo" image is further quantized to obtain (d).}}}
	\label{fig:func}
\end{figure}

We formalize the above discussion as follows. Assume $\mathcal{X} \subseteq \R^{n}$ to be a given (known) subset in the data space, and consider a signal (or image) $x \in \mathcal{X}$. We model (possible) multiplexing operations and gain adjustments as linear transformations, denoted by $A\in\mathbb{R}^{p\times n}$ and $C\in\R^{m\times p}$ respectively. The composite observation model becomes:
\begin{equation}
\label{quan_obs}
u=f(Ax),~y=Q(Cu),
\end{equation}
where $f(\cdot)=\mod(\cdot,R)$ denotes the modulo function with respect to a range parameter $R$ and $Q(\cdot)$ denotes a quantization function. In this paper, we consider a 1-bit quantization function with only two levels, $0$ and $1$. A representative example is shown in Fig.\ \ref{fig:func} where $A$ and $C$ are identity operators. In  Figs~\ref{fig:func}(c)  and~\ref{fig:func}(d), the outputs of the functions $f$ and $Q$ are displayed when a test grayscale image (Fig.\ \ref{fig:func}(a)) is used in the input. Our overall objective is to estimate the original signal $x$ from the set of measurements $y$. 

\subsection{Our contributions}

Clearly, the above estimation procedure is challenging due to the highly non-invertible nature of the observation model. In this paper, we design a systematic approach that takes some initial steps towards resolving this challenge. Our overarching assumption is that the measurement operations $A$ and $C$ are part of the design space. The core idea in our approach is that a very small, but carefully designed, non-adaptive set of measurements can support efficient estimation of the unknown signal.

Our approach follows stagewise. First, we consider the problem of inverting the quantization function, i.e., recovering $u$ from $y = Q(Cu)$. We demonstrate the existence of a linear operator $C$ (together with an efficient reconstruction algorithm) that supports such an inversion. Specifically, our operator $C$ obeys a particular block-diagonal form with weights chosen according to a harmonic progression; see Section~\ref{sec:Model} for details. We only consider 1-bit quantization functions, but similar ideas can presumably be extended for a higher number of quantization levels. In addition, our method supports the criterion of \emph{consistent reconstruction} as defined in \cite{jacques2011dequantizing}.

Next, we consider the problem of inverting the modulo operation, i.e., recovering $x$ from $u = f(Ax)$.  We propose an algorithm that builds upon the approach proposed in \cite{SoltaniHegde_ICASSP16}. In particular, we show that if the operator $A$ satisfies a certain \emph{factorization} $A = DB$, then $f$ can be stably inverted. To enable efficient inversion, the matrix $D$ must also be block-diagonal with weights chosen either randomly, or according to a geometric progression. In the former case, the reconstruction algorithm is an extension of the approach of~\cite{SoltaniHegde_ICASSP16}, while in the latter case the reconstruction follows the approach of~\cite{ICCP15_Zhao}.

The above two-stage procedure can be easily adapted to the case where we have some prior knowledge of the original signal $x$. This enables our approach to be used in conjunction with compressive imaging architectures. Common priors used in compressive imaging include \emph{sparsity} in some known orthonormal basis~\cite{foucart2013}. Note that our measurements are highly quantized and the total ``bit" complexity of our observations is far smaller than conventional techniques. Therefore, within our framework, one can choose to increase the number of quantizer measurements (rows of $C$) and/or modulo measurements (rows of $D$) in order to achieve better estimation performance.

\begin{figure}[t]
	\begin{center}
		\begingroup
		\setlength{\tabcolsep}{1pt} 
		\renewcommand{\arraystretch}{.1} 
		{\setlength{\tabcolsep}{1mm}
		\begin{tabular}{ccc|c|c}      
			\centering
			\includegraphics[trim = 30mm 60mm 40mm 65mm,clip, width = 0.15\linewidth]{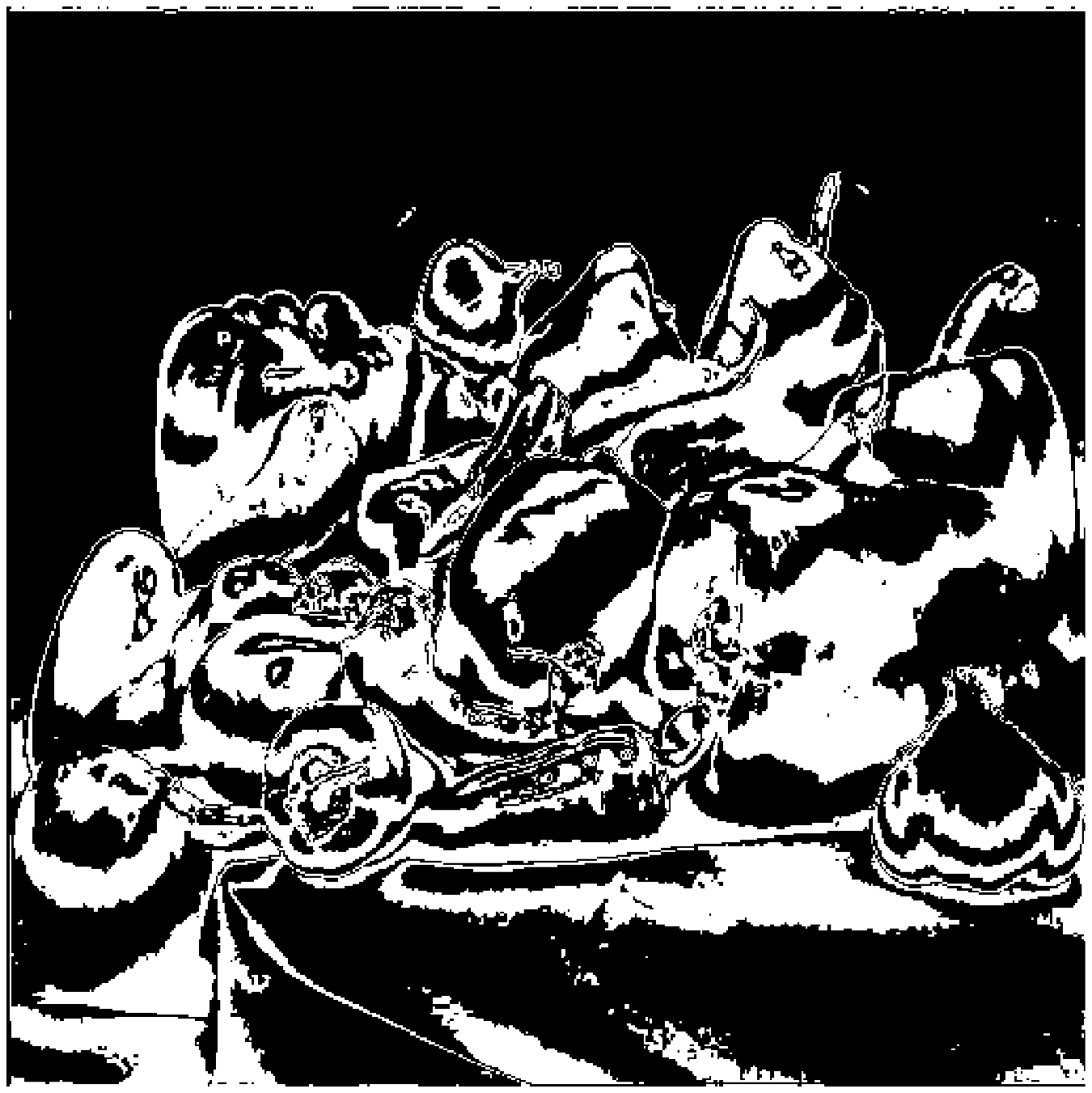}&
			\includegraphics[trim = 30mm 60mm 40mm 65mm,clip, width = 0.15\linewidth]{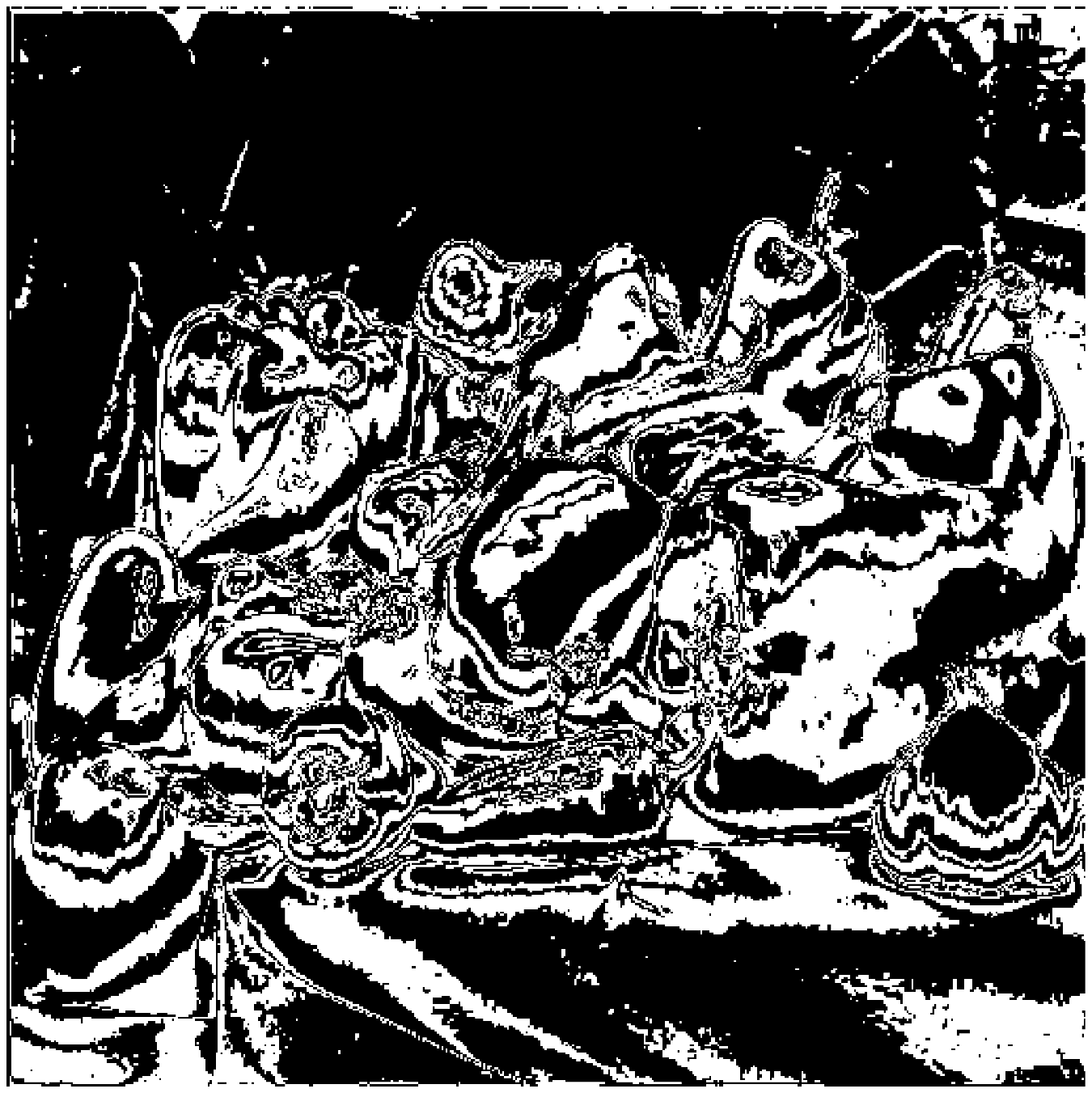}&
			\includegraphics[trim = 30mm 60mm 40mm 65mm,clip, width = 0.15\linewidth]{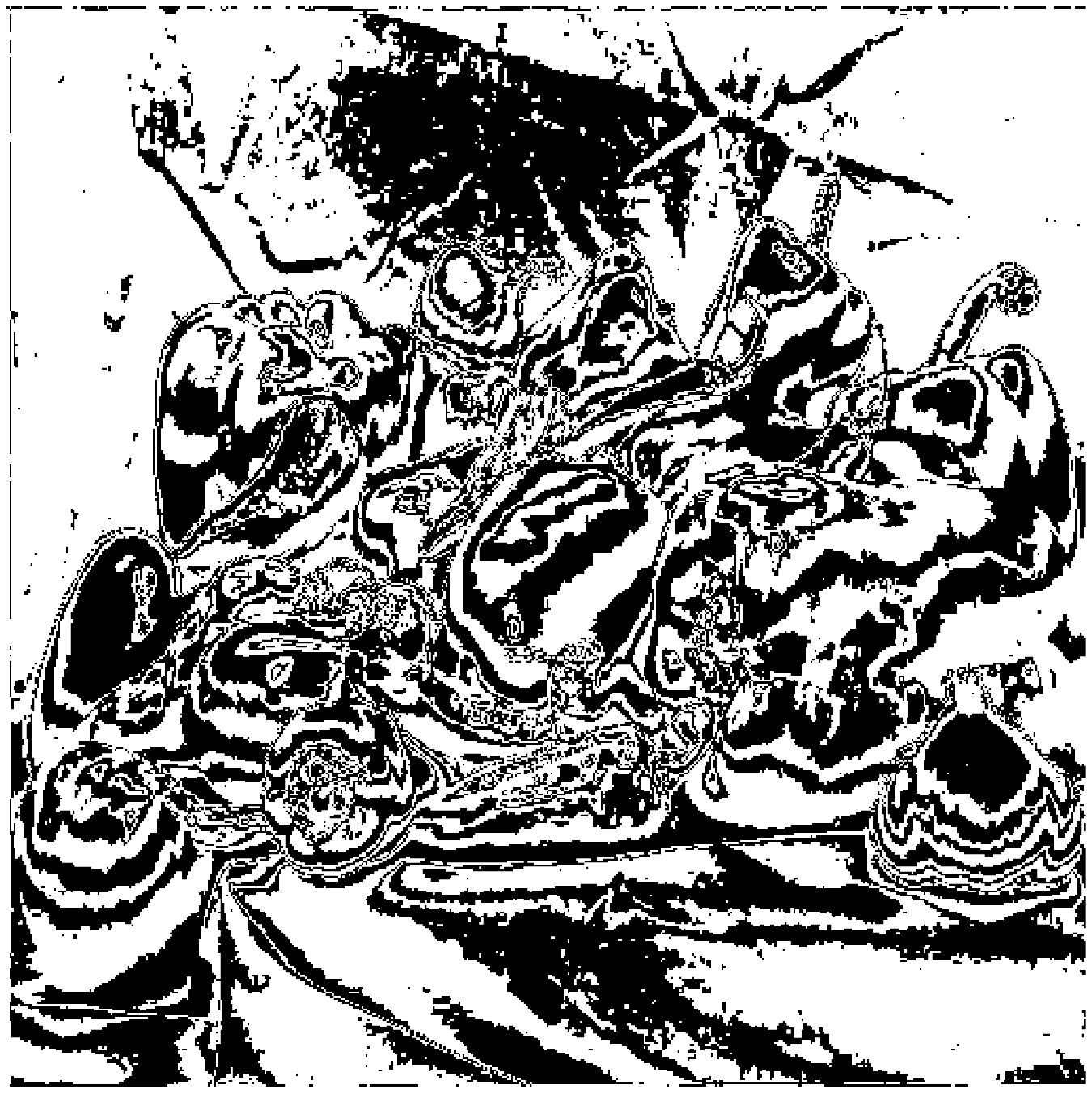}&
			\includegraphics[trim = 90mm 125mm 90mm 120mm,clip, width = 0.18\linewidth]{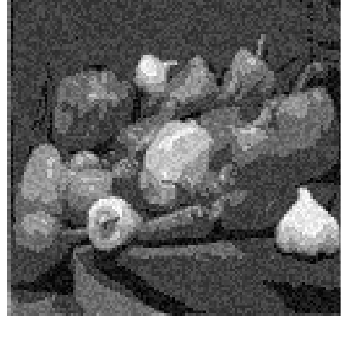}&
				\multirow{3}{20mm}{\includegraphics[trim = 90mm 85mm 90mm 120mm,clip, width = \linewidth]{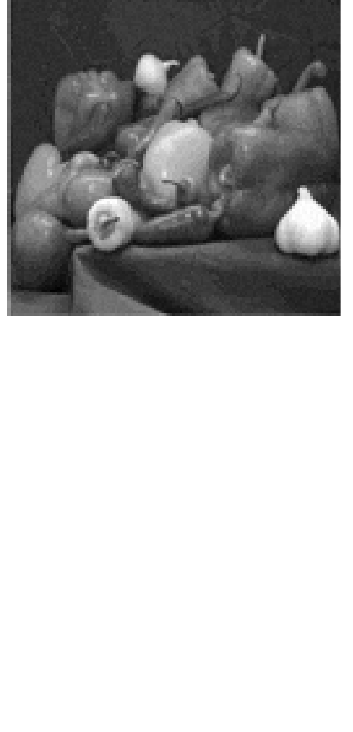}}\\
			\includegraphics[trim = 30mm 60mm 40mm 65mm,clip, width = 0.15\linewidth]{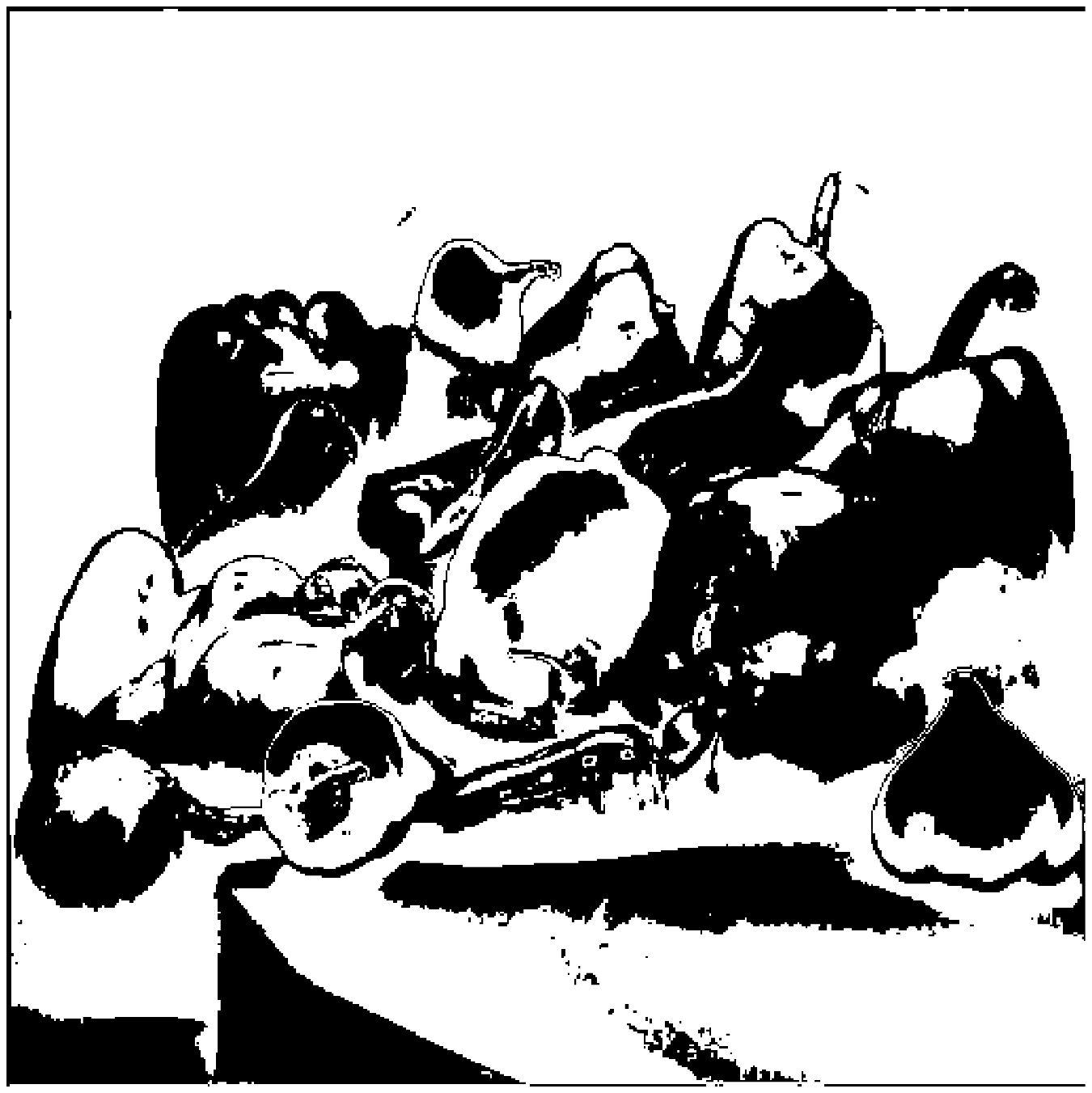}& 
			\includegraphics[trim = 30mm 60mm 40mm 65mm,clip, width = 0.15\linewidth]{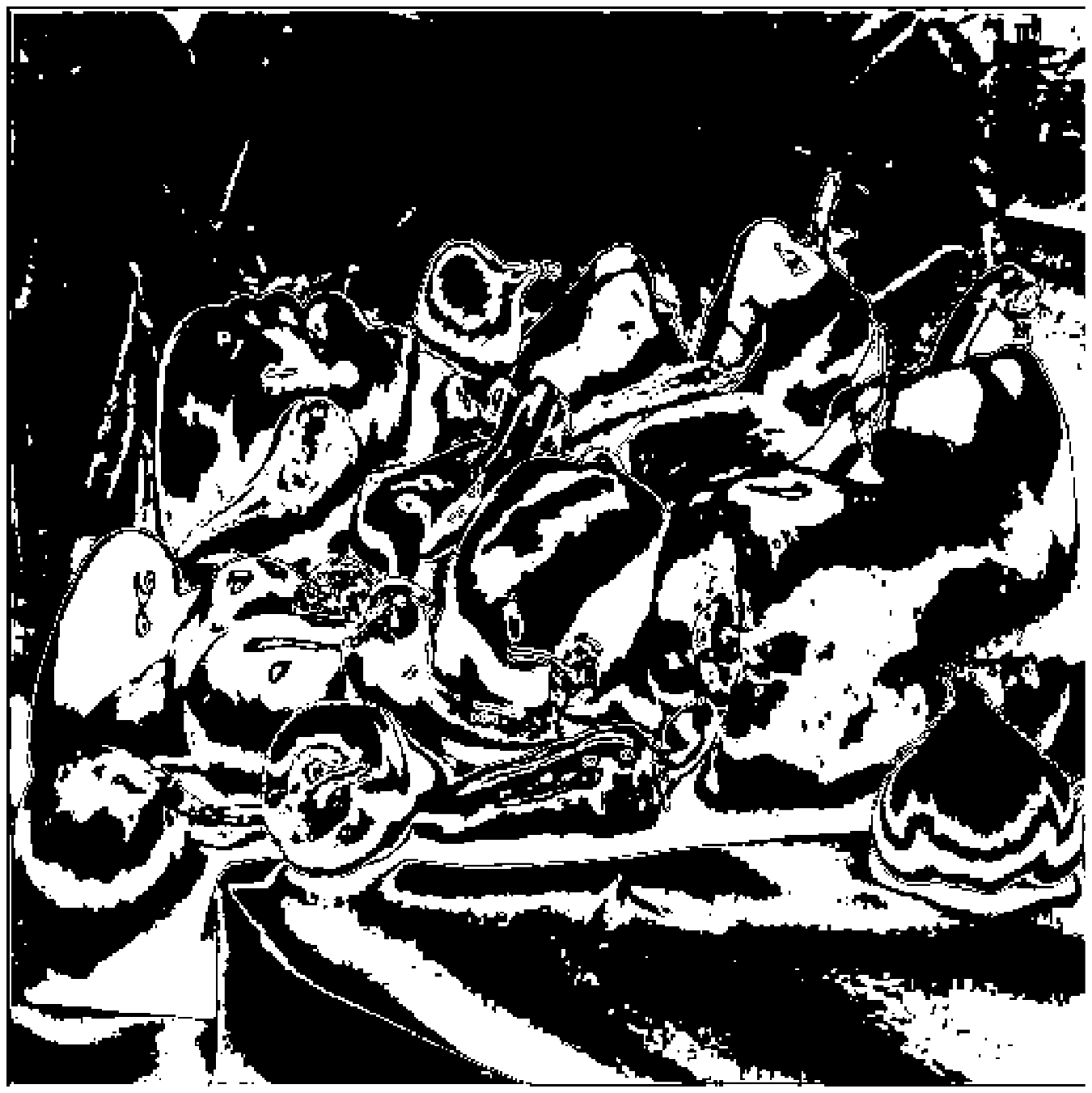}&
			\includegraphics[trim = 30mm 60mm 40mm 65mm,clip, width = 0.15\linewidth]{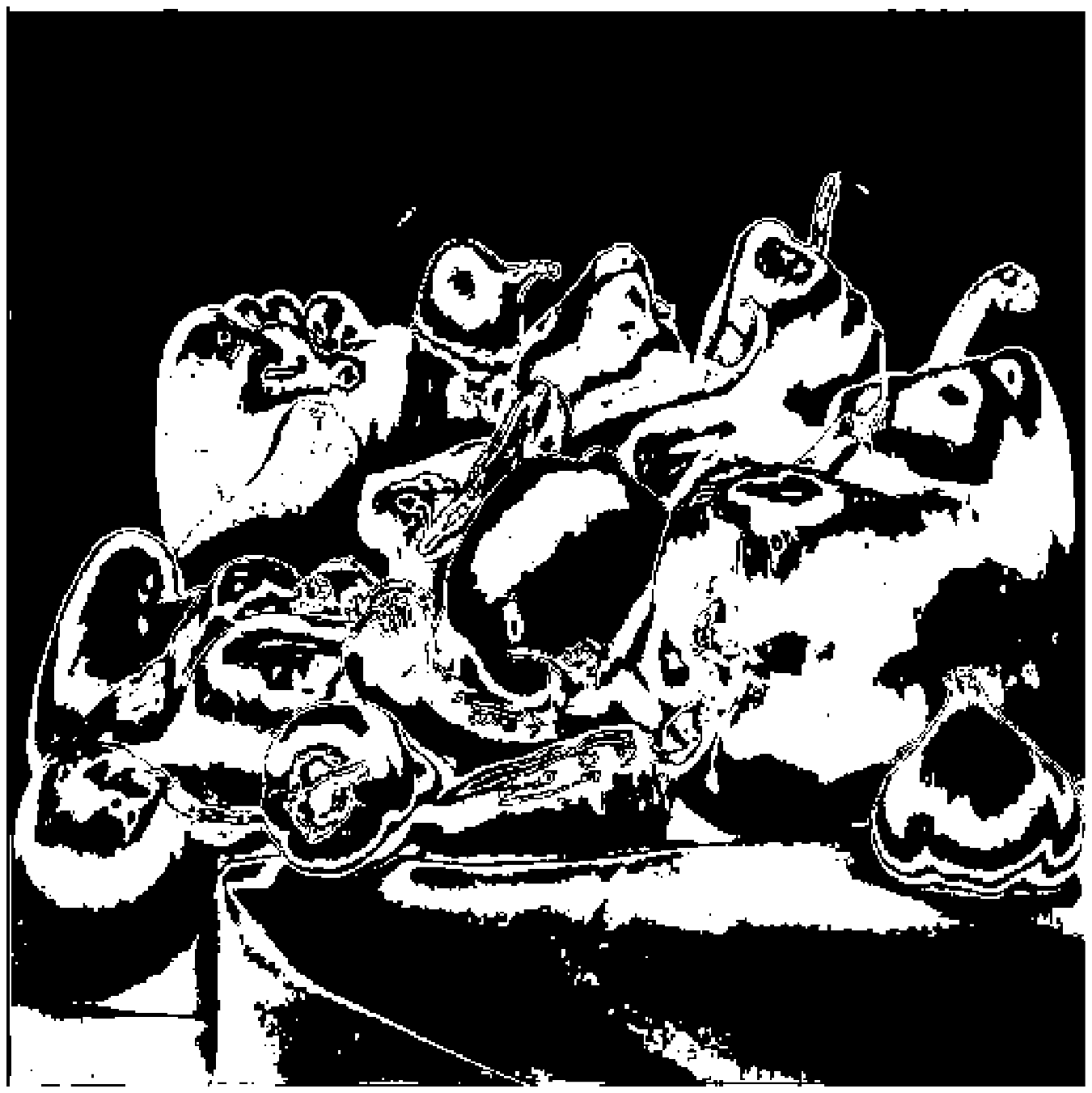}&
			\includegraphics[trim = 90mm 125mm 90mm 120mm,clip, width = 0.18\linewidth]{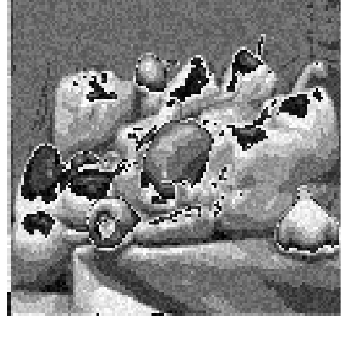}&\\
			\includegraphics[trim = 30mm 50mm 40mm 65mm,clip, width = 0.15\linewidth]{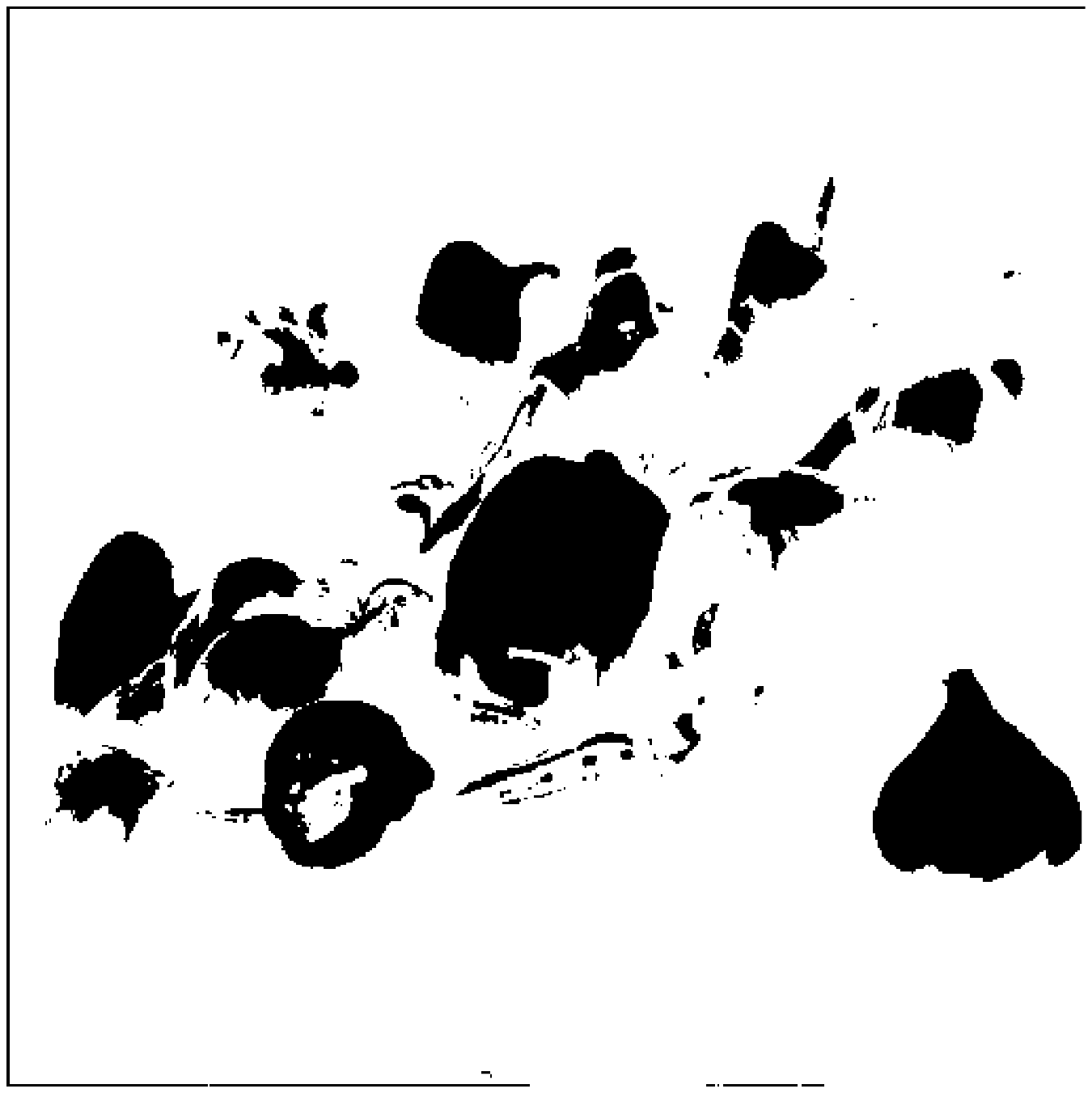}& 
			\includegraphics[trim = 30mm 50mm 40mm 65mm,clip, width = 0.15\linewidth]{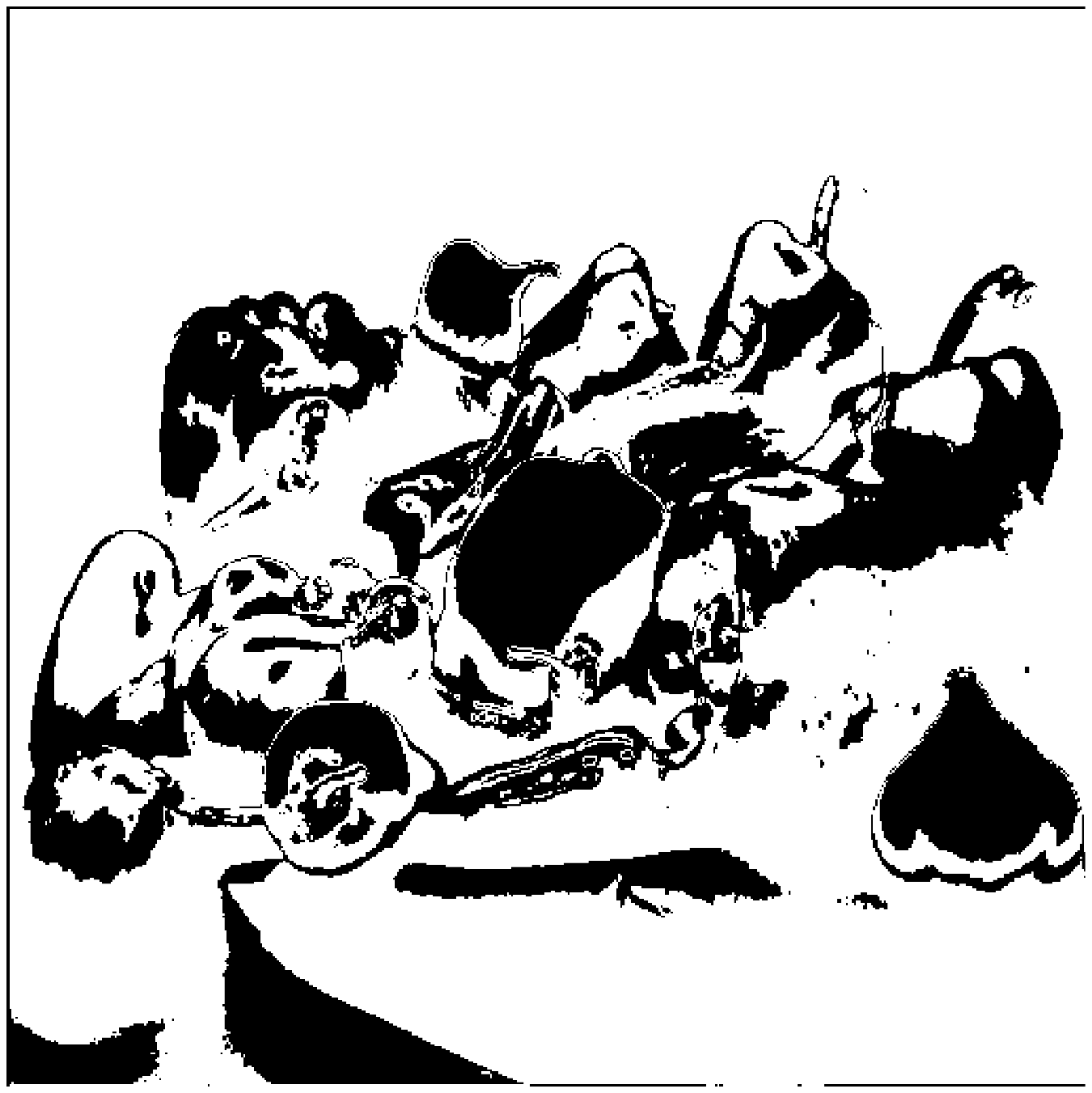}&
			\includegraphics[trim = 30mm 50mm 40mm 65mm,clip, width = 0.15\linewidth]{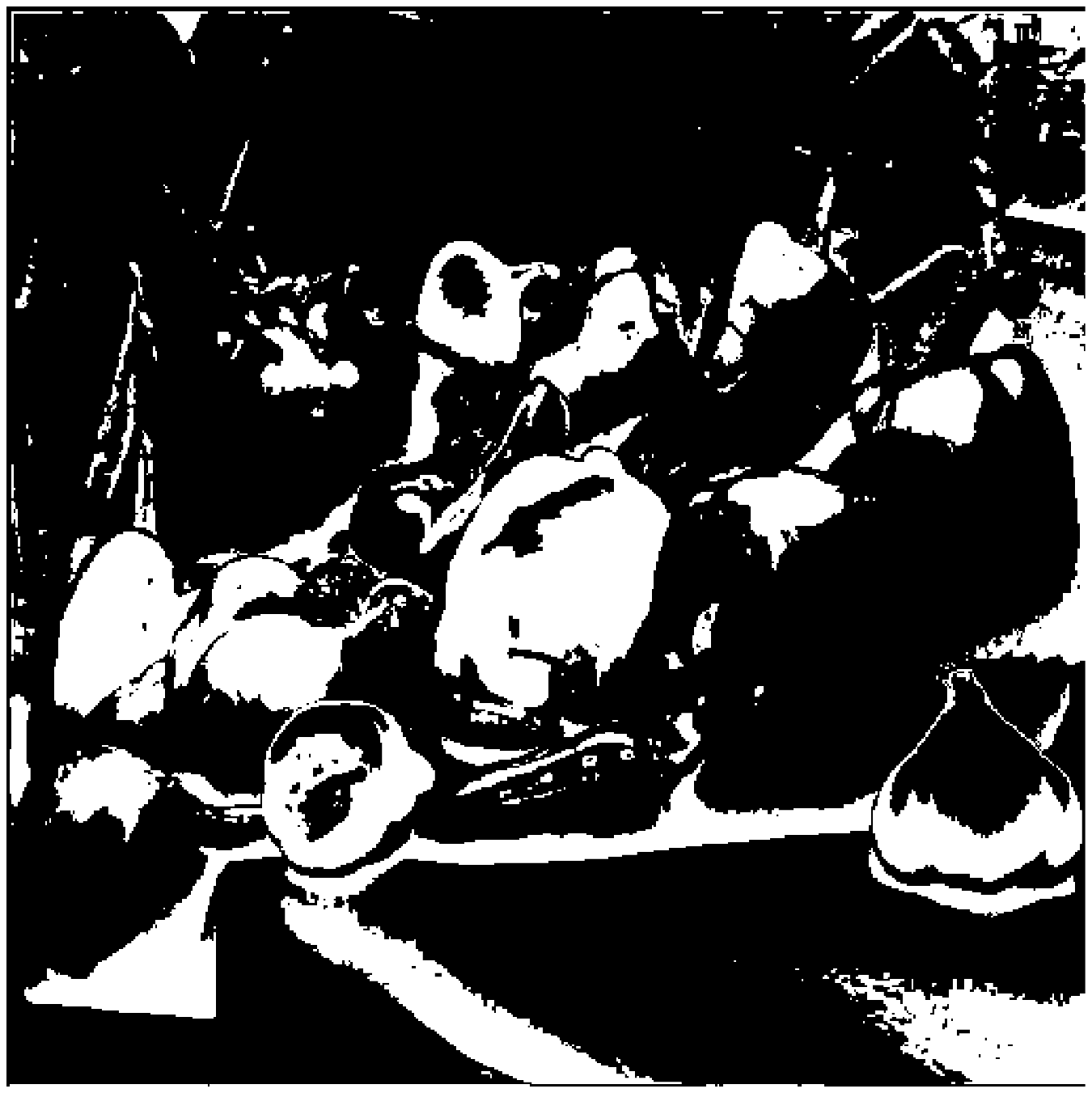}& 
			\includegraphics[trim = 90mm 125mm 90mm 120mm,clip, width = 0.18\linewidth]{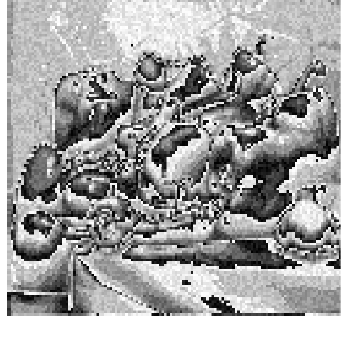}&\\[1pt]
			\multicolumn{3}{c|}{(a)} &(b)&{\centering(c)}
		\end{tabular}}
		\endgroup
	\end{center}
	\caption{\small{\emph{Illustration of our approach. A given input image is modulated pixel-wise with three pre-chosen weights, passed through a modulo sensor, modulated again pixel-wise with three weights, and quantized to binary images. The resulting observations are shown in (a). The images in (b) and (c) represent the reconstruction of the modulo images, $\widehat{u}$ and the final image, $\widehat{x}$, respectively.}}}
	
	\label{fig:demo}
\end{figure}

Fig.~\ref{fig:demo} displays some representative results using our approach. We begin with a standard ``Peppers" image, compute a modulo transformation with three multiplexed measurements per pixel, and further modulate it with a sequence of three harmonic multipliers per measurement before passing it through a 1-bit quantizer. (In words, the overall ``oversampling factor" in our method is $9\times$.) The final binary measurements displayed in Fig.\ \ref{fig:demo}(a) are given as inputs to our reconstruction algorithm. The results from the first and second stages are displayed as images in Fig.\ \ref{fig:demo}(b). As is visually evident, our method is able to successfully reconstruct the image, as displayed in Fig.\ \ref{fig:demo}(c).

\section{Prior Work}
\label{sec:Prior}
\vspace{-0.5em} 
The recovery problem considered in this paper is a confluence of 3 sub-problems --- dequantization, modulo inversion, and sparse recovery. While each of them have been separately considered in considerable detail in signal processing, to the best of our knowledge our work is the first to combine all three sub-problems. The first and third subproblems fall under the purview of quantized compressive sensing and has been the topic of extensive study, dating back to the work of~\cite{DBLP:journals/corr/JacquesC16} and \cite{boufounos20081}. While \cite{DBLP:journals/corr/JacquesC16} uses additive, random dither to compensate the effect of quantization, our method proposes a different, multiplicative approach. Moreover, \cite{boufounos20081} introduces sparse recovery from 1-bit measurements, but strictly assumes that the signal is normalized to have unit Euclidean norm. In \cite{kamilov2012message}, a similar problem of recovering $x$ from $Q(Ax)$ has been studied, but contrary to our setup, it pre-supposes the components of $z=Ax$ to be correlated. 

The modulo inversion subproblem is also known in the literature as \emph{phase unwrapping}. The algorithm proposed in \cite{bioucas2007phase} is specialized to images, and employs graph cuts for phase unwrapping from a single modulo measurement per pixel. However, the inherent assumption there is that the input image has very few sharp discontinuities, and this makes it unsuitable for practical situations with textured images. Our main motivation for this paper is the work of \cite{ICCP15_Zhao} on HDR imaging using a modulo camera sensor. For image reconstruction using multiple measurements, it proposes the method called the multi-shot UHDR recovery algorithm; below, we show that this method can be effectively used in conjunction with quantizers as well as multiplexing mechanisms such as compressive imaging systems. 

Our approach can be viewed as an application of the ``decoupled" measurements idea described in \cite{SoltaniHegde_ICASSP16}. This earlier work did not take into account the effect of quantization in the reconstruction procedure. We resolve this issue using our proposed harmonic multipliers method that we describe in detail below.

\section{Mathematical Model}
\label{sec:Model}
\vspace{-0.2em} 
Let us introduce some notation. Let $x_i$ denote the $i^{\mathrm{th}}$ entry of the vector $x\in\mathbb{R}^n$. Moreover, $x(i:q:(k-1)q+i)$ denotes the sub-vector in $\mathbb{R}^k$ formed by the entries of $x$ starting at index $i + qr$, where $r = 0, 1, \ldots, k-1$ and $i \in [q]$. In addition, $A(i:q:(k-1)q,l)$ represents the sub-vector constructed by selecting the $l^{\mathrm{th}}$ column of matrix $A$ and choosing the entries of the selected column as above.

In~\eqref{quan_obs}, we assume that the matrix $A$ can be factorized as $A=DB$ where $D\in\R^{p\times n}$ is obtained by stacking $k'$ diagonal matrices with size $q\times q$. Moreover, $B\in\R^{q\times n}$ is a matrix which depends on the prior used for $x$. For instance, if $x$ is assumed to be a sparse vector in some orthobasis, then $B$ can be any matrix that supports sparse recovery; for instance, $B$ can satisfy the restricted isometry property (RIP)~\cite{CandesRIP}, or the null-space property (NSP)~\cite{foucart2013}. 

Also, in~\eqref{quan_obs} we assume that $C$ is another block diagonal matrix of size $m\times p$ formed by $k$ diagonal matrices with size $p\times p$ such that the diagonal entries in each of these blocks are specially chosen (see Section~\ref{sec:harmonic} below for more details). We assume that $p$ and $m$ are multiples of $q$ and $p$, respectively. 

For brevity, let us denote $\mod(\cdot,R)$ as $\mod(\cdot)$. Then, the model~\eqref{quan_obs} can be expanded as:
\begin{align}\label{Mainmodel}
y = Q\left(
\begin{bmatrix}
C^0 \\
C^1 \\
\vdots  \\
C^{k-1}
\end{bmatrix}
\mod\left(
\begin{bmatrix}
D^1 \\
D^2 \\
 \vdots  \\
D^{k'}
\end{bmatrix}
Bx
\right)\right) .
\end{align}

We define the following intermediate variables:
\begin{align}\label{mainabr}
u = \mod(DBx)\in\R^p, \ \ z = Bx\in\R^q,
\end{align}
As discussed in~\cite{SoltaniHegde_ICASSP16}, the block diagonal structure of $D$ and $C$ allows the signal reconstruction problem to be reduced to a sequence of \emph{decoupled} scalar estimation problems; such a decoupling enables the estimation of each entry of $z$ and $u$ independently of all other entries.

We also assume that the analog values of the input signal lies within a range known \emph{a priori}. 
We define $\Delta$ as the reference point for the quantization, and assume that the inputs to the quantizer are bounded within the region $[0,2\Delta]$. 
The function under consideration is the 1-bit quantization function, 
defined as follows:
\begin{align}\label{eq:quantfunc}
Q(u_i)= 
\begin{cases}
0,& \text{if } u_i\leq {\Delta}\\
1,              & \text{if } {\Delta} < u_i \leq {2\Delta} 
\end{cases}
\end{align}

For every element of the input to the quantizer, $u_i$, we measure multiple outputs $y_{i,0},y_{i,1}, \cdots ,y_{i,k-1}$, given by:
\begin{align}\label{eq:hmshort}
y_{i,j} = Q(c_{i,j}u_i) , ~~~~ j = 0,1,2,...,k-1.
\end{align}
with $c_0 =1$, and each subsequent $c_{i,j}$ is defined as :
\begin{align}\label{eq:hm}
c_{i,j} = 
\begin{cases}
\frac{k}{k-j},& \text{if } y_{i,0} = 0,\\
\frac{k}{k+j},              & \text{if } y_{i,0} = 1,
\end{cases} ~~~~ j = 1,2,...,k-1.
\end{align}
The underlying idea is to increase or decrease the value of $c_{i,j}u_i$ gradually and to detect the index $j^*$ for which $y_{i,j}$ changes its value for the first time. Using $j^*$, Interval containing $u_i$ can be determined. Here, the reason for choosing the values of $c_{i,j}$ in harmonic progression is to ensure that all such intervals corresponding to the different values of $j^*$ have equal widths. This fact is made clearer in Section~\ref{sec:harmonic}.

As the multipliers $c_{i,j}$ form a harmonic progression for both the cases, we call the proposed measurement scheme the \emph{harmonic multipliers method}. In order to express it as a linear transformation, these multipliers can be arranged into a block diagonal matrix $C$ of size $(kp) \times p$, for which the diagonal entry in $i^{th}$ row in block $C^j$ will be $c_{i,j}$, the multiplier corresponding to the input signal element $u_i$. 
The forward model can be written in the form of following equation:
\begin{align}
y = Q(Cu),
\end{align}
where $u$ is defined in~\eqref{mainabr}.

\section{Reconstruction Procedure}
To solve the inverse problem in~\eqref{Mainmodel}, we propose a three stage procedure that we call \textit{reconstruction from de-quantized modulo observations}, or \textit{RQM} for short. In the first stage, RQM estimates $u= mod(Ax)$ from vector $y$. Next, it uses the estimate, $\widehat{u}$ to produce an estimate $z$ in the second stage (say $\widehat{z}$). Finally, this estimate $\widehat{z}$ is used for recovery of the original $x$. The pseudocode of RQM is given as Alg.~\ref{alg:DMF}.  We now describe each of these stages in detail. 

\begin{algorithm}[t]
\caption{\textsc{RQM}}
\label{alg:DMF}
\begin{algorithmic}
\State\textbf{Inputs:} $y$, $D$, $B$, $C$, $k$, $k'$, $\Omega$, $s$
\State\textbf{Output:}  $\widehat{x}$
\State \textbf{Stage 1: Harmonic dequantization}
\State $\widehat{u}\leftarrow \textsc{HMDequantization}(y,C,k)$
\State \textbf{Stage 2: Modulo recovery}
\State $\theta\leftarrow  \exp(i \widehat{u})$
\For {$l =1:q$}
\State $t \leftarrow D(l:q:(k'-1)q+l,l)$
\State $\phi \leftarrow \theta(l:q:(k'-1)q+l)$
\State $\widehat{z_l} = \argmax_{\omega\in\Omega}|\langle y,\psi_{\omega}\rangle|$
\EndFor
\State $\widehat{z} \leftarrow [\widehat{z_1},\widehat{z_2}\ldots,\widehat{z_q}]^T$
\State \textbf{Stage 3: Sparse recovery}
\State $\widehat{x} \leftarrow \textsc{CoSaMP}(\widehat{z},B,s)$
\end{algorithmic}
\end{algorithm}

\subsection{Harmonic dequantization}
\label{sec:harmonic}
We first attempt to recover $u$. Based on the value of $Q(u_i)$, we can know the interval in which $u_i$ lies, which can either be $[0,\Delta]$ or $[\Delta,2\Delta]$. For $u_i \in [0,\Delta]$, the multipliers are designed in such a way that with each multiplication, we gradually increase the value of $c_{i,j}u_i$ until it becomes greater than $\Delta$ at $j=j^*$ to give $Q(c_{i,j^*}u_i)=1$. 

Using $j^*$, we can decide the interval of $u_i$ as follows. 
Equation \eqref{eq:hmshort} gives:
\[
y_{i,{j^*-1}} = Q(c_{i,j^*-1} u_i) = \left \lfloor{\dfrac{k u_i}{(k-j^*+1) \Delta}}\right \rfloor = 0.
\]
Similarly, 
\[
y_{i,{j^*}} = Q(c_{i,j^*} u_i) = \left \lfloor{\dfrac{k u_i}{(k-j^*) \Delta}}\right \rfloor = 1.
\]
Combining the above relations, we infer that:
\begin{align}
\label{eq:hminter1}
\Delta \dfrac{(k-j^*)}{k} \leq u_i < \Delta \dfrac{(k-j^*+1)}{k}.
\end{align}
Similarly, for $u_i$ with $y_{i,0}=1$, through each multiplication with $c_{i,j}$, the value of $c_{i,j} u_i$ decreases gradually and becomes less than $\Delta$ for $j=j^*$ for the first time. 
\begin{align}
\label{eq:hminter2}
\Delta \dfrac{(k+j^*-1)}{k} \leq u_i < \Delta \dfrac{(k+j^*)}{k}.
\end{align}
\begin{algorithm}[t]
	\caption{\textsc{HMDequantization}}
	\label{alg:HM}
	\begin{algorithmic}
		\State\textbf{Inputs:} $y$, $C$, $k$
		\State\textbf{Output:}  $\widehat{u}$\\
		$n \leftarrow length(y)/k$
		\For {$l =1:n$}
		\If {$y_l = 0$}
		\State $t \leftarrow y(l+n:n:(k-1)n+l,1)$
		\State $j^* \leftarrow \min_{j \in \{1,2,...,k-1\}} \text{such that } t_j = 1$
		\State $\widehat{u}_{l} \leftarrow v \sim U[\Delta\frac{k-j^*}{k},\Delta
		\frac{k-j^*+1}{k}]$
		\ElsIf {$y_l = 1$}
		\State $t \leftarrow y(l+n:n:(k-1)n+l,1)$
		\State $j^* \leftarrow \min_{j \in \{1,2,...,k-1\}} \text{such that } t_j = 0$
		\State $\widehat{u}_{l} \leftarrow v \sim U[\Delta\frac{k+j^*-1}{k},\Delta\frac{k+j^*}{k}]$
		\EndIf
		\EndFor
		
	\end{algorithmic}
\end{algorithm}
Equations \eqref{eq:hminter1} and \eqref{eq:hminter2} provide us the interval on the real line that contains $u_i$. To remove bias, a random real number is chosen from this interval as the final estimate $\widehat{u_i}$. The width of this interval is $\delta = \frac{\Delta}{k}$, which is same for every interval corresponding to different values of $j^*$ owing to harmonic design of the multipliers. The value of $\delta$ (and consequently, the estimation error) can be made sufficiently small by increasing the value of $k$. For the estimate $\widehat{u_i}$ to lie within an $\epsilon \Delta$ neighborhood of $u_i$, the minimum value required for $k$ can be calculated as 
$k_{\text{req}} = \left \lceil{\frac{1}{\epsilon}}\right \rceil$.

In this paper, we assume that the first multiplier $c_0 =1$, and it is possible to decide the appropriate $c_j$'s by looking at the value of the first measurement $y_{i,0}$ in Eq\ \eqref{eq:hm}. 
In case purely non-adaptive measurements are desired, a similar approach can be followed by acquiring $2k-1$ measurements with each possible value of $c_j$ specified by both cases of~\eqref{eq:hm}. The pseudocode of this stage is given as Alg~\ref{alg:HM}.

\vspace{-1.2em}
\subsection{Modulo recovery}\label{ToneEst}
	\vspace{-0.5em}
The output of \textsc{HMDequantization} acts as input for the modulo recovery stage. The goal of this stage is to find an estimate for the vector $z$. There are several different ways of doing this, including the multi-shot UHDR method of~\cite{ICCP15_Zhao}. Here, we describe a novel approach, based on the MF-Sparse algorithm of~\cite{SoltaniHegde_ICASSP16}. We assume that the entries of $z$ belong to some bounded set $\Omega \in R$. Fix $l \in [q]$ and form $\theta =  \exp(i \widehat{u})$. Let {$t = D(l:q:(k'-1)q+l,l)$ and $\phi = \theta(l:q:(k'-1)q+l)$}, 
which are vectors in $\mathbb{R}^{k'}$. Thus, we have the following model:
\begin{align}\label{tonmodel}
\phi = \exp(i z_l t) \, .
\end{align}
In the above model, $\phi$ can be interpreted as a set of time samples of a complex-valued signal with frequencies $z_l \in \Omega$, measured at time locations $t$. As a result, we can independently recover $z_l$ for $l=1, \ldots, q$ by solving a least-squares problem~\cite{eftekhari2013matched}:
\begin{align}
\label{OptmExp}
\widehat{z_l} = \underset{v \in \Omega}{\argmin}~\|\phi - \exp(i \, vt )\|_2^2 = \underset{v \in \Omega}{\argmax}~\left|\langle \phi,\psi_{v}\rangle\right|,
 \end{align}
for all $l=1,\ldots,q$, where  $\psi_{v}\in\mathbb{R}^{k'}$ denotes a \emph{template vector} given by $\psi_{v} = \exp(j t v)$ for any $v \in \Omega$. The solution of this optimization problem is equivalent to performing a \emph{matched filter} from irregularly spaced samples. Numerically, the optimization problem in~\eqref{OptmExp} can be solved using a grid search over the set $\Omega$, and the resolution of this grid search controls the running time of the algorithm. For fine enough resolutions, the estimation of $z_l$ is more accurate, at the expense of increased running time. This issue is also discussed in~\cite{eftekhari2013matched} and~\cite{eldarxampling,tangcsoffgrid,chioffgrid} have proposed more sophisticated spectral estimation techniques. 

In~\eqref{tonmodel}, the vector $\theta$ is modeled in terms of complex exponentials. As discussed in~\cite{SoltaniHegde_ICASSP16}, we can equivalently use a real-valued sine function. That is, the vector $\phi$ can be defined as:
\begin{align}
\label{eq:realsine_obs}
\phi = \sin(i z_l t),
\end{align}
Similar to the complex case, we estimate $z$ by solving~\cite{SoltaniHegde_ICASSP16}:
\begin{align*}
\label{OptmSin}
\widehat{z_l} &= \underset{v \in\Omega}{\argmin}~\|\phi - \sin(v t)\|_2^2 = \underset{v \in \Omega}{\argmax}~\left( 2\left|\langle \phi,\psi_{v}\rangle\right| - \|\psi_{v}\|_2^2\right),
\end{align*}
for $l=1,\ldots,q$ and $\phi$ as defined above and $\psi_v =\sin(tv)$. 
	\vspace{-1.2em}
\subsection{Sparse recovery}
	\vspace{-0.5em}
Finally, we estimate the original signal $x$ from $\widehat{z}$ obtained as the output of the second stage.
Note that the use of sparse recovery here is generic, and we could in principle use any other prior model of relevance to the specific imaging application. Since we assume that matrix $B$ in~\eqref{Mainmodel} supports stable sparse recovery and the underlying signal $x$ is $s$-sparse, we can use any generic sparse recovery algorithm to estimate $x$. In our experiments, we chose to use the CoSaMP algorithm \cite{cosamp} due to its ease and speed. Hence, we take $\widehat{z}$ from previous stage and run CoSaMP to obtain the final estimation, $\widehat{x}$.

\section{Experimental Results}
\label{sec:Results}
\begin{figure*}[t]
	\begin{center}
		\begingroup
		\setlength{\tabcolsep}{1pt} 
		\renewcommand{\arraystretch}{.1} 
		\begin{tabular}{ccc}      
			\includegraphics[trim = 30mm 80mm 35mm 80mm, clip, width=0.25\linewidth]{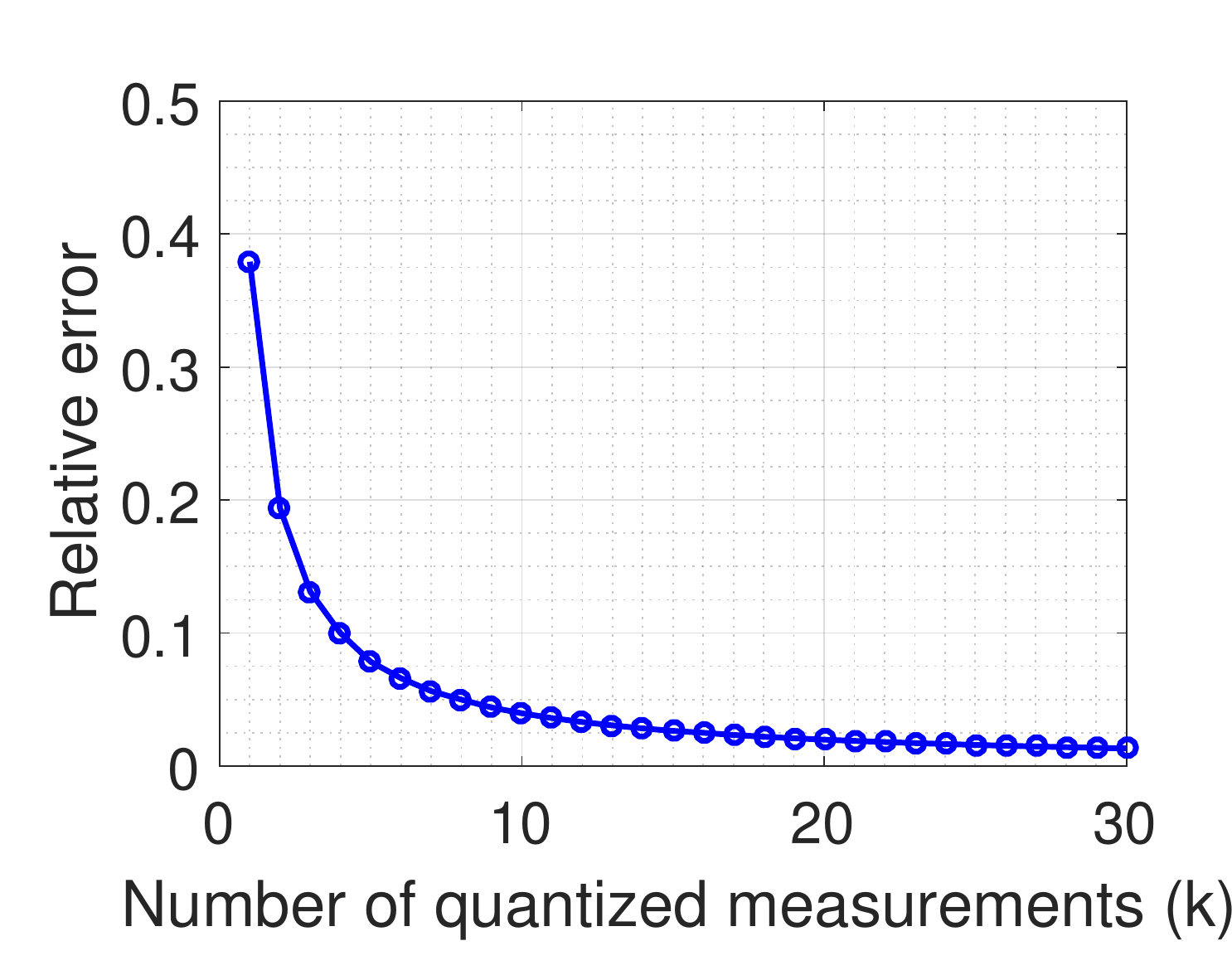}&
			\includegraphics[trim = 30mm 80mm 35mm 80mm, clip, width=0.25\linewidth]{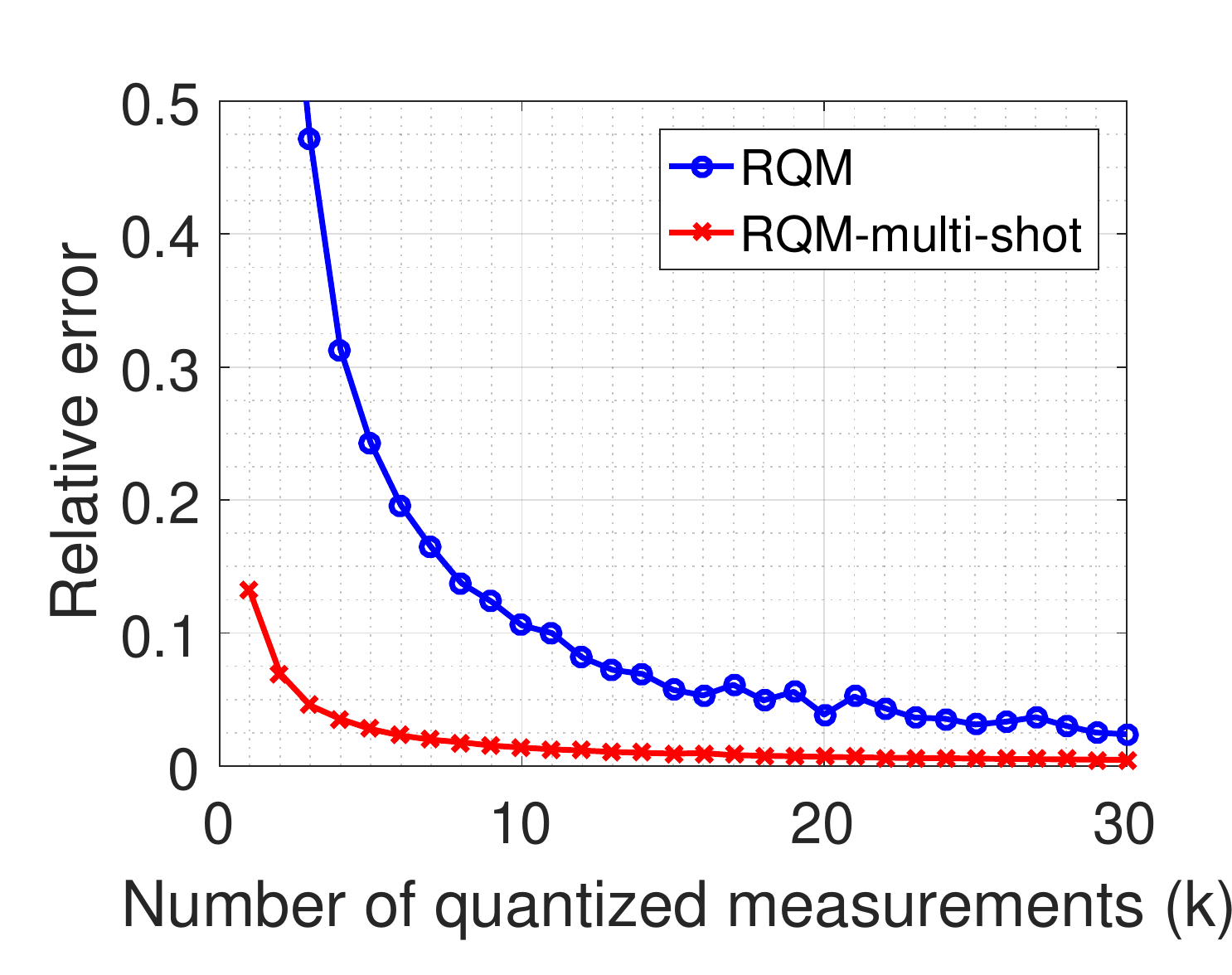}&
			\includegraphics[trim = 30mm 80mm 35mm 80mm, clip, width=0.25\linewidth]{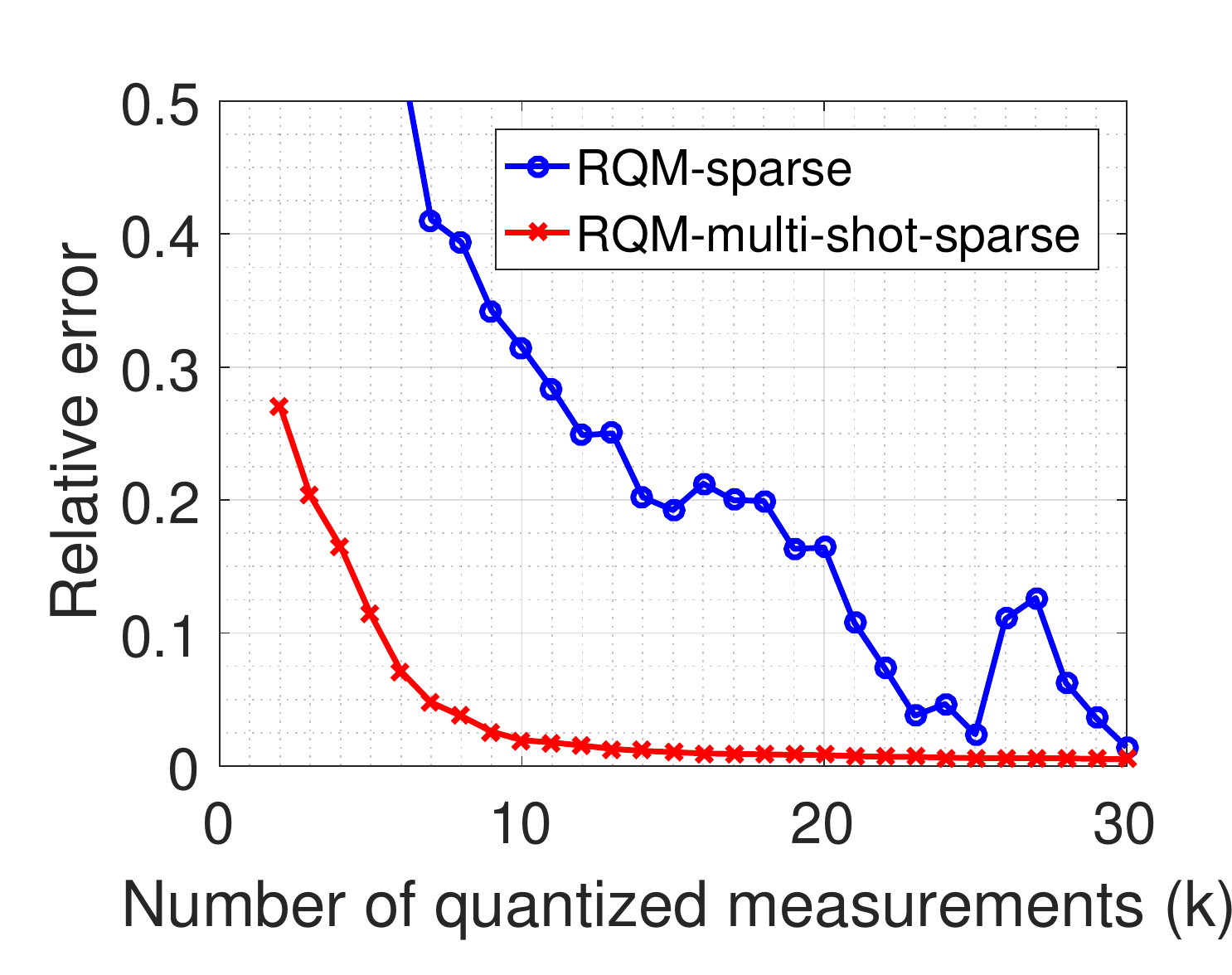}
			\\
			(a) & (b) & (c)
		\end{tabular}
		\endgroup
	\end{center}
	\caption{\emph{ Normalized error vs number of quantized measurements $(k)$ for: (a) dequantization using HM algorithm; (b) reconstruction from quantized modulo measurements using RQM and RQM multi-shot; (c) reconstruction from quantized modulo measurements of sparse input using RQM and RQM multi-shot.}}
	\label{fig:results}
\end{figure*}

\begin{figure*}[t]
	\begin{center}
		\begingroup
		\setlength{\tabcolsep}{0.1pt} 
		\begin{tabular}{cccccc}      

				\includegraphics[trim = 35mm 75mm 45mm 77mm, clip, width=0.12\linewidth]{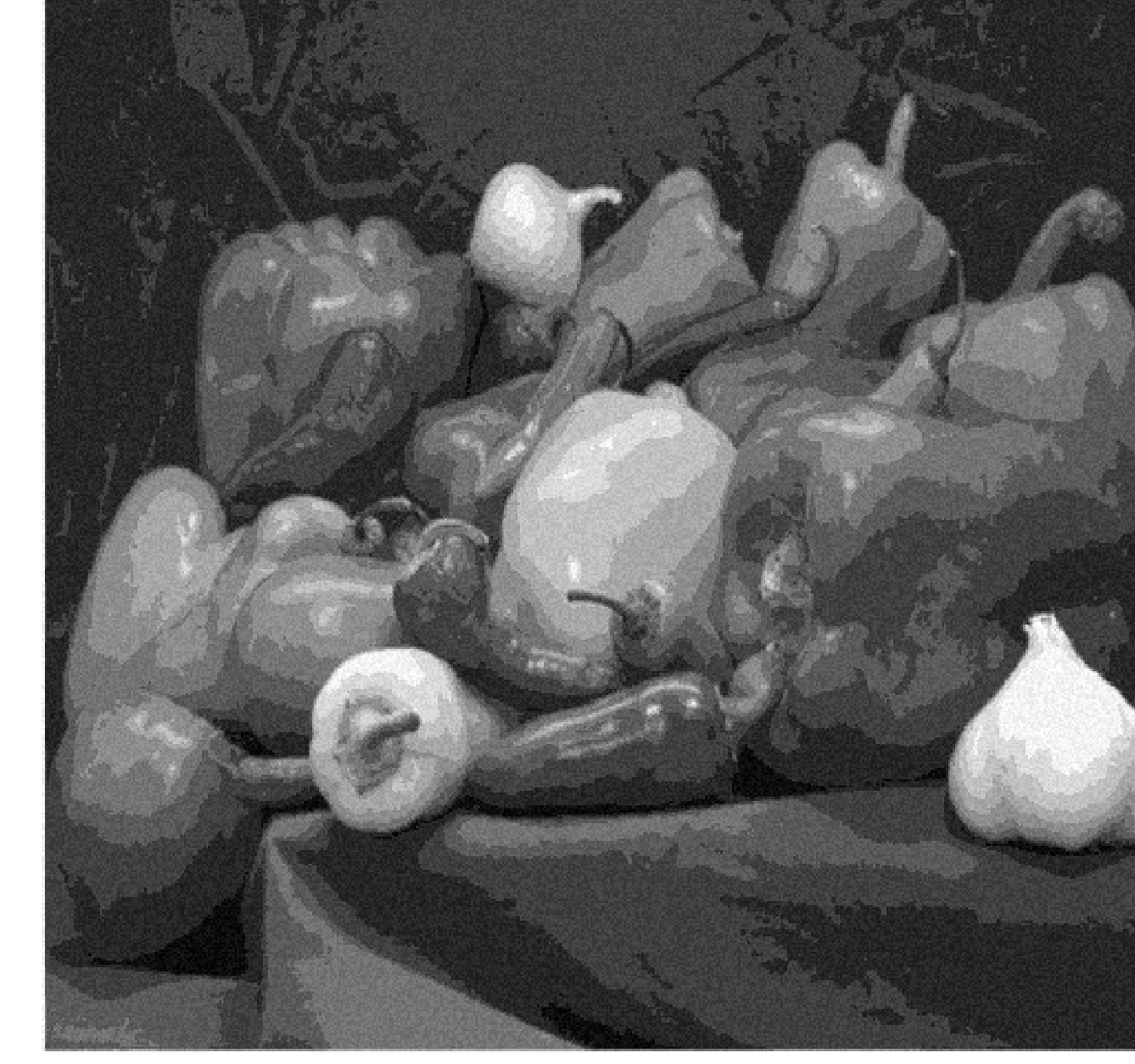}& \hspace{15pt}
				\includegraphics[trim = 89mm 127.25mm 92mm 119.15mm, clip, width=0.12\linewidth]{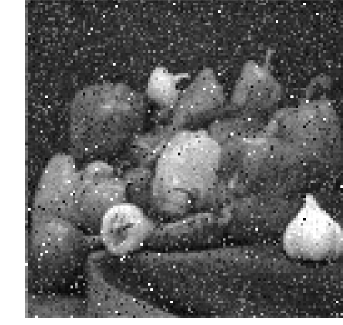}&
					\includegraphics[trim = 89mm 126.5mm 92mm 120mm, clip, width=0.12\linewidth]{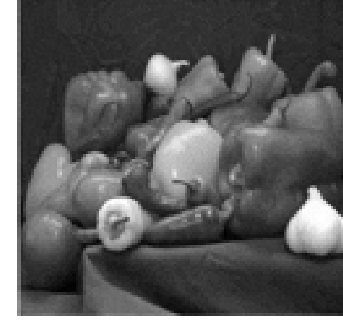}& \hspace{15pt}
						\includegraphics[trim = 70mm 109mm 75mm 103.5mm, clip, width=0.12\linewidth]{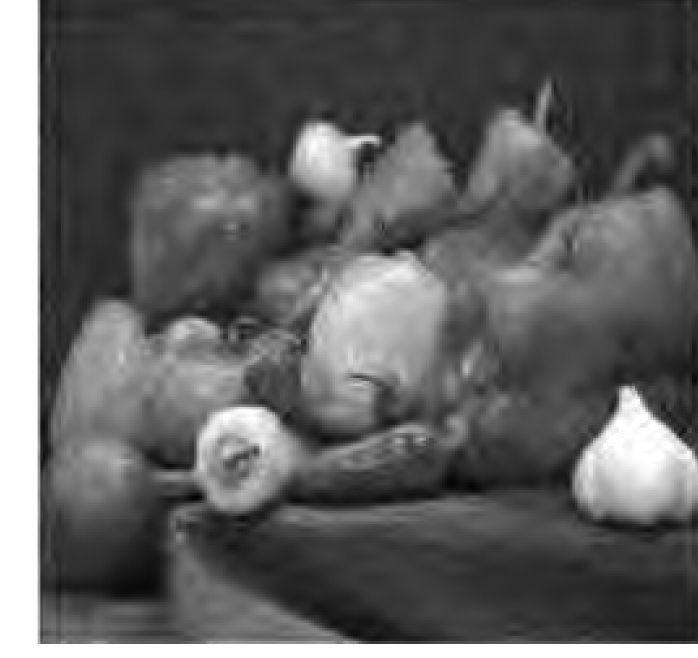}&
							\includegraphics[trim = 70mm 109mm 75mm 103mm, clip, width=0.12\linewidth]{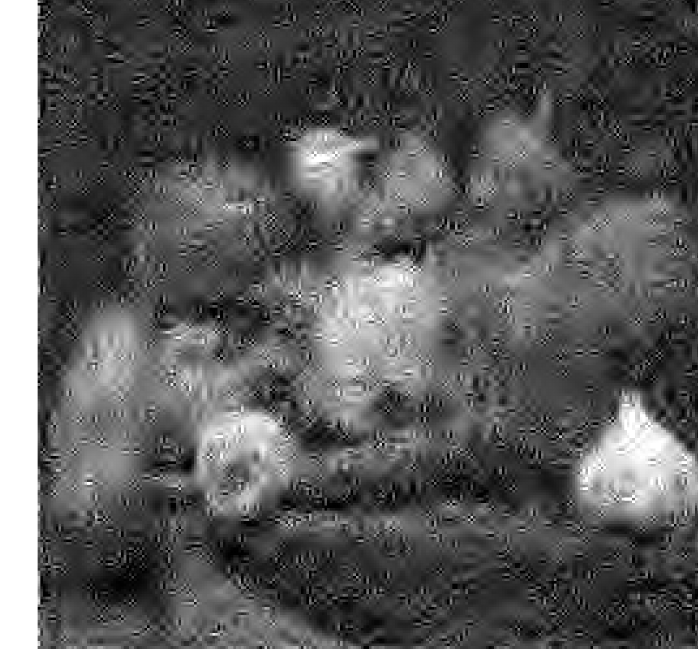}&
								\includegraphics[trim = 70mm 110mm 75mm 100.5mm, clip, width=0.12\linewidth]{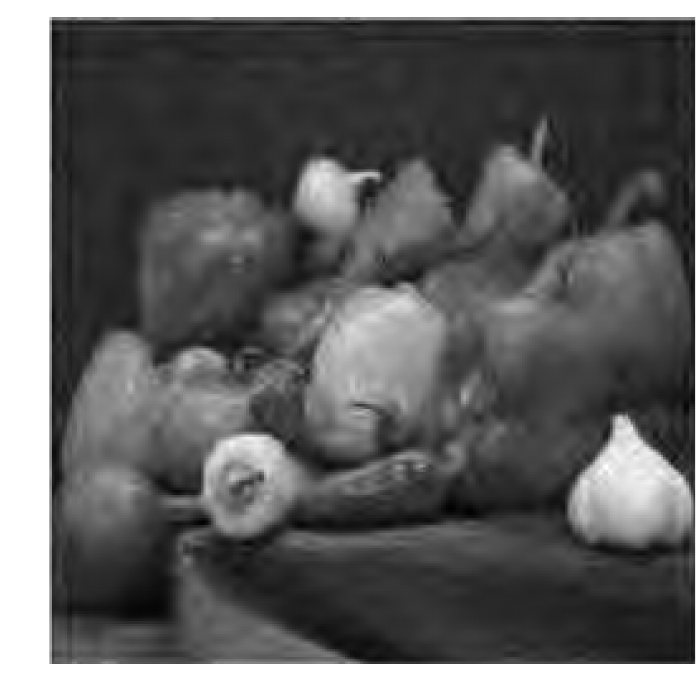} \\

			 (a) & \multicolumn{2}{c}{\hspace{20pt}(b)} & \multicolumn{3}{c}{\hspace{22pt}(c)}
		\end{tabular}
		\endgroup
	\end{center}
	\caption{\emph{Image reconstruction results: (a) image reconstructed from quantized measurements using HM, $k=5$; (b) image reconstructed from quantized modulo measurements using RQM (left) and RQM multi-shot (right), $k=5$; (c) sparse input image (left), image reconstructed from quantized modulo measurements of sparse input using RQM-sparse (centre) and RQM multi-shot-sparse (right), $k=10$.}}
	\vspace{-0.4em}
	\label{fig:imgresult}
\end{figure*}
	\vspace{-0.5em}
In this section, we provide some representative numerical experiments for our proposed algorithm. 
First, we provide results describing our proposed dequantization procedure on a real test image. Further, we also provide results for the combined task of de-quantization and modulo recovery, with and without sparsity priors on the underlying signal. We employ two different algorithms for modulo recovery, and therefore we have following four combinations for our experiments: (i) reconstruction from de-quantized modulo observations using RQM, with and without sparsity priors, (ii) de-quantization using our HM algorithm followed by modulo recovery using the multi-shot UHDR recovery algorithm (we refer this whole procedure as \emph{RQM-multi-shot}), with and without sparsity priors.
	\vspace{-0.7em}
\subsection{Dequantization}
	\vspace{-0.4em}
In this experiment, we only focus on the first stage, i.e., we attempt to recover the image $w$ from set of $k$ quantized measurements $y^j$, $j=0,1,2,...,k-1$, using the \textsc{HMDequantization} method for different values of $k$. We record the normalized estimation error defined as $\frac{\|\widehat{w}-w\|_2}{\|w\|_2}$, with $\widehat{w}$ being the estimate of $w$. Here, $w$ is the grayscale form of an 8-bit, 3-channel RGB image of the size $512 \times 512$. The 1-bit quantizer described in Eq \eqref{eq:quantfunc} is used with $\Delta = 2^7$ to calculate $y$. Based on values of $y^0$, the coefficients $c_j$s are decided for each element of $w$. Subsequently, $(k-1)$ measurements are obtained according to Eq \eqref{eq:hm}, and the \textsc{HMDequantization} method is used to obtain $\widehat{w}$. The normalized estimation error is plotted against the number of measurements $k$ in Fig.~\ref{fig:results}(a). As we observe from the plot, our algorithm can recover $w$ within $10\%$ of error with as low as $5$ measurements. Increasing the value of $k$ improves the recovery performance rapidly in this regime, and less than $5\%$ error can be achieved with $k\geq9$. 

	\vspace{-0.7em}
\subsection{Experiment: No sparsity priors}
	\vspace{-0.3em}
We take an 8-bit, 3-channel RGB image of size $256 \times 256$, convert it to grayscale, and scale the dynamic range to [0,1]. Since there are no sparsity priors assumed here, we let $B$ be the identity matrix. We consider two cases for $D$. In the forward model specified by the RQM algorithm, the vectorized image $x$ is first multiplied by the block diagonal matrix $D_{mf}$. The size of $D_{mf}$ is set $(k'n) \times n$ as it contains $k'$ blocks of size $n \times n$ each. Diagonal of each block contains uniformly distributed random variables in the range $[-T,T]$.  Similarly, in the forward model specified by for RQM-multi-shot, $x$ is multiplied by the block diagonal matrix $D_{ms}$; here, all diagonal elements for $r^{th}$ block are same and equal to $2^{9-r}$; for $r = 1,2, \ldots, k'$ \cite{ICCP15_Zhao}. 

To recover $\widehat{z} = \widehat{x}$, the estimation of $z =x$, from measurement $y$, we employ both the RQM as well as the RQM-multi-shot algorithms in two separate experiments.
In Fig.~\ref{fig:results}(b), we plot the normalized estimation error in recovered $x$ in case of RQM-multi-shot by varying $k$, while $k'$ is fixed to 3. As we can see, we are able to recover the original image within $5\%$ error only with $k=3$ quantized measurements. Fig.~\ref{fig:results}(b) also shows the variation of normalized estimation error for the RQM algorithm with $k'$ fixed to 4. To recover the original image within $5\%$ error, RQM requires $k=15$ quantized measurements. 

	\vspace{-0.7em}
\subsection{Experiment: Sparsity priors}
	\vspace{-0.3em}
We now evaluate the performance of the proposed method in scenarios where the input signal is $s$-sparse. We use the same $256 \times 256$ RGB image, convert to grayscale, and after obtaining its 2D Haar wavelet decomposition, retain the $s=1000$ largest coefficients to sparsify the image. We further multiply the sparse test image by a subsampled Fourier matrix with $q=8000$ multiplied with a diagonal matrix with random $\pm1$ entries to get $z=Bx$. The rest of the observation process is identical to the experiment described above.

Again, two separate experiments are performed with using RQM in one and RQM-multi-shot algorithm in another to recover $\widehat{z}$ from $y$. The final step is to compute the estimate of high dimensional signal $\widehat{x} \in \mathbb{R}^n$ from $\widehat{z} \in \mathbb{R}^q$, which we achieve using the CoSamP algorithm~\cite{cosamp}.
For the RQM-multi-shot-sparse algorithm, we fix $k' = 3$, and obtain the plot of relative error by varying the value of $k$, which is shown in Fig.~\ref{fig:results}(c). As we can see from the plot, we are able to recover the original image within $5\%$ error only with the use of $7$ quantized measurements. Similar to the experiment without sparsity, we fixed $k'=4$ for RQM-sparse algorithm, and measure the normalized estimation error in $\hat{x}$ for different values of $k$. Corresponding plot is in Fig.~\ref{fig:results}(c). To estimate the original image within $5\%$ relative error, $k=25$ quantized measurements are used.

Considering that the input is sparse and the measurements $y$ are binary, the storage requirements for $y$ are considerably smaller compared to the case without sparsity. In essence, leveraging the sparsity prior can reduce the sample complexity of the algorithm drastically. The tradeoff is to choose a higher value of $k$ or $k'$, which will affect the running time only marginally but improves recovery performance by a significant amount.


{{
\footnotesize
\bibliographystyle{IEEEbib}
\bibliography{../Common/chinbiblio,../Common/csbib,../Common/mrsbiblio,../Common/vsbib,../Common/kernels}
}
}

\end{document}